\documentclass[letterpaper]{article}
\usepackage{aaai16}
\usepackage{times}
\usepackage{helvet}
\usepackage{courier}
\title{A Generative Model of Words and Relationships from Multiple Sources}
\author{Stephanie L. Hyland\textsuperscript{a,b} \and  Theofanis
Karaletsos\textsuperscript{a} \and Gunnar R\"{a}tsch\textsuperscript{a} \\
\textsuperscript{a} Computational Biology Program, Memorial Sloan Kettering
Cancer Center\\ 1275 York Avenue, New York, NY 10065\\\textsuperscript{b}
Tri-Institutional Training Program in Computational Biology and Medicine\\
Weill Cornell Medical College, 1305 York Ave, New York, NY
10021\\\texttt{\{stephanie, theo, gunnar\}@ratschlab.org}}
\usepackage{url}
\usepackage{graphicx}
\usepackage{amsmath}
\usepackage{amssymb}
\usepackage{algorithm}
\usepackage{color}
\newcommand{\citetext}[1]{\citeauthor{#1} \shortcite{#1}}

\begin{document} \maketitle
%%%%

\begin{abstract}
Neural language models are a powerful tool to embed words into semantic vector 
spaces. However, learning such models generally relies on the availability of 
abundant and diverse training examples. In highly specialised domains this
requirement may not be met due to difficulties in obtaining a large corpus,
or the limited range of expression in average use. Such domains may encode prior
knowledge about entities in a knowledge base or ontology. We propose a
generative model which integrates evidence from diverse data sources, enabling
the sharing of semantic information. We achieve this by generalising the concept
of co-occurrence from distributional semantics to include other relationships
between entities or words, which we model as affine transformations on the
embedding space. We demonstrate the effectiveness of this approach by
outperforming recent models on a link prediction task and demonstrating its
ability to profit from partially or fully unobserved data training labels. We
further demonstrate the usefulness of learning from different data sources with
overlapping vocabularies.
\end{abstract}

\section{Introduction\footnote{A preliminary version of this work appeared at the International Workshop on Embeddings and Semantics at SEPLN 2015 \cite{hylandIWES}.}}

A deep problem in natural language processing is to model the semantic
relatedness of words, drawing on evidence from text and spoken language, as
well as knowledge graphs such as ontologies. A successful modelling approach
is to obtain an \emph{embedding} of words into a metric space such that
semantic relatedness is reflected by closeness in this space. One paradigm
for obtaining this embedding is the \emph{neural language model} \cite{Bengio:2003:NPL:944919.944966}, which traditionally draws on local co-occurence
statistics from sequences of words (sentences) to obtain an encoding of words
as vectors in a space whose geometry respects linguistic and semantic features.
The core concept behind this procedure is the \emph{distributional hypothesis
of language}; see \citetext{sahlgren2008distributional}, that semantics can be
inferred by examining the \emph{context} of a word. This relies on the
availability of a large corpus of diverse sentences, such that a word's typical
context can be accurately estimated.

In the age of web-scale data, there is abundant training data available for
such models in the case of generic language. For \emph{specialised} language
domains this may not be true. For example, medical text data \cite{liu-EtAl:2015:BioNLP15}
often contains protected health information, necessitating access restrictions
and potentially limiting corpus size to that obtainable from a single 
institution, resulting in a corpus with less than tens of millions of sentences,
not billions as in (for example) Google n-grams. In addition to this, specialised domains
expect certain prior knowledge from the reader. A doctor may never mention that
anastrazole  is a aromatase inhibitor (a type of cancer drug), for example, 
because they communicate sparsely, assuming the reader shares their training
in this terminology. In such cases, it is likely that even larger quantities of 
data are required, but the sensitive nature of such data makes this difficult to
attain.

\begin{figure} \includegraphics[scale=0.25]{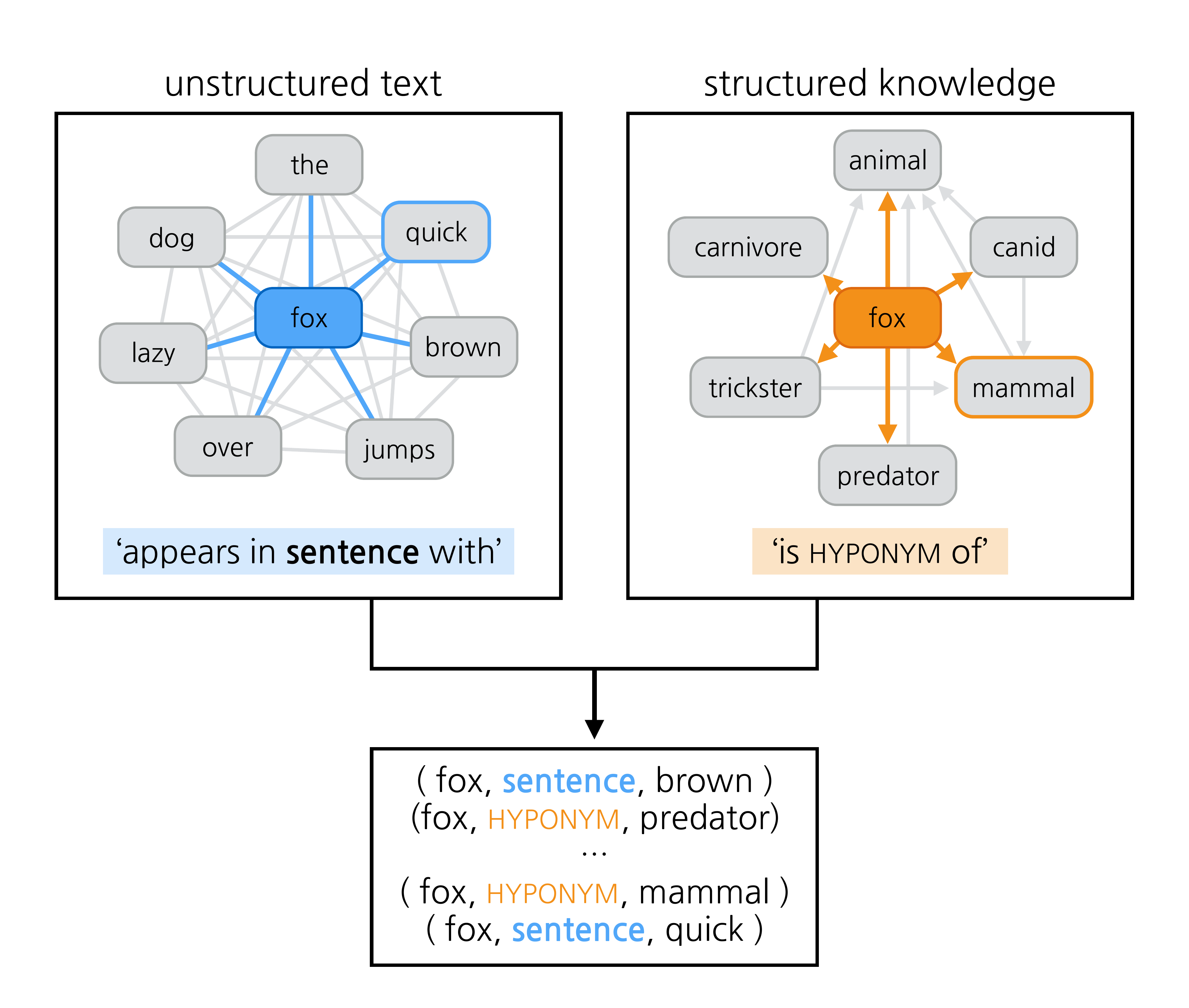} \caption{We
unify structured and unstructured data sources by considering functional (e.g.
hyponymic) relationships to be a form of \emph{co-occurrence}, and considering
sentence co-occurrence to be another type of functional relationship. Thus, our
model is source-agnostic and uses true $(S, R, T)$ triples as evidence to
obtain an embedding of entities and relationships.\vspace{-3ex}}
\label{fig:illustration} \end{figure}

Fortunately, such specialised disciplines often create expressive
\emph{ontologies}, in the form of annotated relationships between terms (denoted by underlines). These
may be semantic, such as \underline{dog} is a \underline{type of}
\underline{animal}, or derived from domain-specific knowledge, such as
\underline{anemia} is an \underline{associated disease of}
\underline{leukemia}. (This is a relationship found in the medical ontology
system UMLS; see \citeauthor{bodenreider2004unified}, \citeyear{bodenreider2004unified}). We observe that these
relationships can be thought of as additional \emph{contexts} from which
co-occurrence statistics can be drawn; the set of diseases associated with
leukemia arguably share a common context, even if they may not co-occur in
a sentence (see \textbf{Figure~\ref{fig:illustration}}).

We would like to use this structured information to improve the quality of
learned embeddings, to use their information content to regularize the
embedding space in cases of low data abundance while obtaining an explicit
representation of these relationships in a vector space. We tackle this by 
assuming that each relationship is an \emph{operator} which transforms words in
a relationship-specific way. Intuitively, the action of these operators is to 
distort the shape of the space, effectively allowing words to have multiple 
representations without requiring a full set of parameters for each possible 
sense.

The intended effect on the underlying (untransformed) embedding is twofold: to
encourage a solution which is more sensitive to the domain than would be
achieved using only unstructured information and to use heterogeneous sources
of information to compensate for sparsity of data. In addition to this, since
relationship operators can act on \emph{any} part of the space, by learning 
these functions we can apply them to any word regardless of its source,
allowing for link prediction on new entities in a knowledge base.

While we do not attempt to model higher-order language structure such as
grammar and syntax, we consider a generative model in which the distance
between terms in the embedded space describes the probability of their
co-occurrence under a given relationship. Through this, we learn the joint
distribution of all pairs in all relationships, and can ask questions such as
\emph{`What is the probability of \underline{anemia} appearing \underline{in a 
sentence with} \underline{imatinib}\footnote{Imatinib is a tyrosine-kinase inhibitor 
used in the treatment of chronic myelogenous leukemia.}?'}, or \emph{`What is the
probability that \underline{anemia} is a \underline{disease associated} with
\underline{leukemia}?'} This introduces flexibility for subsequent analyses that
require a generative approach.

This paper is laid out as follows: In Related Work, we describe relevant prior
work concerning embedding words and relationships and place our contributions in
context. In Modelling, we describe in detail the probabilistic model and
our inference and learning strategy, including a link to code. In Experiments, we
show an array of experimental results to quantitatively demonstrate different
aspects of the model on datasets using WordNet and Wikipedia sources in
supervised, semi-supervised and unsupervised settings, before summarising our
findings in the Discussion section.

\section{Related Work}
The task of finding continuous representation for elements of language has been
explored in great detail in recent and less-recent years. \citetext{Bengio:2003:NPL:944919.944966} described a neural architecture to predict
the next word in a sequence, using distributed representations to overcome the
curse of dimensionality. Since then, much work has been devoted to obtaining, 
understanding, and applying these distributed language representations. One such model is \texttt{word2vec} of 
\citetext{mikolov2013distributed}, which more explicitly relies on the 
distributional hypothesis of semantics by attempting to predict the surrounding 
context of a word, either as a set of neighbouring words (the skip-gram model) 
or as an average of its environment (continuous bag of words). We note later in
the model section that the idealised version of skip-gram \texttt{word2vec} is
a special case of our model with one relationship; \underline{appears in a sentence with}. In practice, \texttt{word2vec} uses a 
distinct objective function, replacing the full softmax with an 
approximation intended to avoid computing a normalising factor. We retain a
probabilistic interpretation by approximating gradients of the partition
function, allowing us to follow the true model gradient while maintaining
tractability. Furthermore, learning a joint distribution facilitates imputation
and generation of data, dealing with missing data and making predictions
using the model itself. We note that a generative approach to language was also
explored by \citetext{andreas2013generative}, but does not concern relationships.

Relational data can also be used to learn distributed representations of 
entities in knowledge graphs, entities which may correspond to or can be mapped
to words. A general approach is to implicitly embed the graph structure through
vertex embeddings and rules (or transformations) for traversing it. 
\citetext{bordes2011learning} scored the similarity of entities under a given 
relationship by their distance after transformation using pairs of 
relationship-specific matrices. \citetext{socher2013reasoning} describe
a neural network architecture with a more complex scoring function, noting that
the previous method does not allow for interactions between entities. The
\texttt{TransE} model of \citetext{bordes2013translating} (and
extensions such as \citetext{wang2014knowledge}, \citetext{fan2014transition}, and \citetext{lin2015learning})
represents relationships as \emph{translations}, motivated by the tree
representation of hierarchical relationships, and observations that linear
composition of entities appears to preserve semantic meaning
\cite{mikolov2013distributed}. These approaches are uniquely concerned with
relational data however, and do not consider distributional semantics from free
text. \citetext{faruqui2015retrofitting} and \citetext{johansson-nietopina:2015:NAACL-HLT}
describe methods to modify pre-existing word embeddings to align them with 
evidence derived from a knowledge base, although their models do not learn
representations \emph{de novo}.

Similar in spirit to our work is \citetext{weston2013connecting}, where entities
belonging to a structured database are identified in unstructured (free) text
in order to obtain embeddings useful for relation prediction. However, they
learn separate scoring functions for each data source. This approach is also
employed by \citetext{fried2015incorporating}, \citetext{xu2014rc}, \citetext{yu2014improving}, and \citetext{wang2014knowledgeb}. In
these cases, separate objectives are used to incorporate different data
sources, combining (in the case of \citetext{xu2014rc}) the skip-gram objective
from \citetext{mikolov2013distributed} and the \texttt{TransE} objective of
\citetext{bordes2013translating}. Our method uses a single energy function over the
joint space of word pairs with relationships, combining the `distributional
objective' with that of relational data by considering free-text co-occurrences
as another type of relationship.

We have mentioned several approaches to integrating graphs into embedding
procedures. While these graphs have been derived from knowledge bases or
ontologies, other forms of graphs have also been exploited in
related efforts, for example using constituency trees to obtain sentence-level
embeddings \cite{socherlstm2015}.

The motivation for our work is similar in spirit to multitask and transfer
learning (for instance, \citetext{Caruana1997}, \citetext{Evgeniou2004}, or
\citetext{widmer2012multitask}). In transfer learning one takes advantage of data
related to a similar, typically supervised, learning task with the aim of
improving the accuracy of a specific learning task. In our case, we have the
unsupervised learning task of embedding words and relationships into a vector
space and would like to use data from another task to improve the learned
embeddings, here word co-occurrence relationships. This may be understood as
a case of \emph{unsupervised transfer learning}, which we tackle using
a principled generative model.

Finally, we note that a recent extension of \texttt{word2vec} to full
sentences~\cite{jernitefast} using a fast generative model exceeds the scope of
our model in terms of sentence modeling, but does not explicitly model latent
relationships or tackle transfer learning from heterogeneous data sources.

\section{Probabilistic Modelling of Words and Relationships}

We consider a probability distribution over triplets $(S,R,T)$ where $S$ is the
\emph{source word} of the (possibly directional) \emph{relationship} $R$ and
$T$ is the \emph{target word}. Note that while we refer to `words', $S$ and $T$
could represent any entity between which a relationship may hold without
altering our mathematical formulation, and so could refer to multiple-word 
entities (such as \texttt{UMLS} Concept Unique Identifiers) or even non-lexical
objects. Without loss of generality, we proceed to refer to them as words. 
Following \citetext{mikolov2013distributed}, we learn two representations for each
word: $\mathbf{c}_{s}$ represents word $s$ when it appears as a \emph{source}, 
and $\mathbf{v}_{t}$ for word $t$ appearing as a \emph{target}.\footnote{\citetext{goldberg2014word2vec} provide a motivation for using two representations for each word. 
We can extend this by observing that words with similar $\mathbf{v}$ 
representations share a \emph{paradigmatic} relationship in that they may be 
exchangeable in sentences, but do not tend to co-occur. Conversely, words $s$ 
and $t$ with $\mathbf{c}_{s}\approx\mathbf{v}_{t}$ have a \emph{syntagmatic} 
relationship and tend to co-occur (e.g. \citetext{sahlgren2008distributional}).  
Thus, we seek to enforce syntagmatic relationships and through transitivity 
obtain paradigmatic relationships of $\mathbf{v}$ vectors.  } Relationships act
by altering $\mathbf{c}_{s}$ through their action on the vector space
($\mathbf{c}_{s} \mapsto G_{R}\mathbf{c}_{s}$). By allowing $G_{R}$ to be an
arbitrary affine transformation, we combine the bilinear form of
\citetext{socher2013reasoning} with translation operators of
\citetext{bordes2013translating}. 

The joint model is given by a Boltzmann probability density function,
\begin{eqnarray} 
P(S,R,T|\Theta) &= \frac{1}{Z(\Theta)}e^{-\mathcal{E}(S,R,T|\Theta)} \nonumber\\
&=\frac{e^{-\mathcal{E}(S,R,T|\Theta)}}{\sum_{s,r,t}e^{-\mathcal{E}(s,r,t|\Theta)}}
\end{eqnarray}
Here, the partition function is the normalisation factor over the joint posterior 
space captured by the model parameters 
$Z(\Theta)=\sum_{s,r,t}e^{-\mathcal{E}(s, r, t | \Theta)}$.  The parameters
$\Theta$ in this case are the representations of all words (both as sources and
targets) and relationship matrices; $\Theta =\{ \mathbf{c}_i, G_{r}, \mathbf{v}_j,
\}_{i, j \in \textrm{vocabulary}}^{r \in \textrm{relationships}}$. If we
choose an energy function \begin{equation} \mathcal{E}(S,R,T|\Theta)
= -\mathbf{v}_T\cdot G_R \mathbf{c}_S \label{eqn:dotenergy} \end{equation} we
observe that the $|R|=1$, $G_{R} = \mathbb{I}$ case recovers the original
softmax objective described in \citetext{mikolov2013distributed}, so the idealised \texttt{word2vec}
 model is a special case of our model.

This energy function is problematic however, as it can be trivially minimised 
by making the norms of all vectors
tend to infinity. While the partition function provides a global regularizer,
we find that it is not sufficient to avoid norm growth during training. We
therefore use as our energy function the negative cosine similarity, which does
not suffer this weakness;\footnote{We also considered an alternate, more symmetric energy function using the Frobenius norm of $G$; \[\mathcal{E}(S,R,T|\Theta) = -\frac{\mathbf{v}_T\cdot G_R \mathbf{c}_S}{\|\mathbf{v}_T\| \|G_R{\|}_F \|\mathbf{c}_S\|}\] However, we found no clear empirical advantage to this choice.}
\begin{equation} \mathcal{E}(S,R,T|\Theta) = -\frac{\mathbf{v}_T\cdot G_R
\mathbf{c}_S}{\|\mathbf{v}_T\|\| G_R \mathbf{c}_S\|} \end{equation} This is
also a natural choice, as cosine similarity is the standard method for
evaluating word vector similarities. Energy minimisation is therefore equal to
finding an embedding in which the \emph{angle} between related entities is
minimised in an appropriately transformed relational space. It would be simple to define a more complex energy function (using perhaps splines) by choosing a different functional representation for $R$, but we focus in this work on the affine case.

\paragraph{Inference and Learning} We estimate our parameters $\Theta$ from data
using stochastic maximum likelihood on the joint probability distribution. The 
maximum likelihood estimator is: 
\begin{equation} \Theta^* = \text{argmax}\;P(\mathcal{D}|\Theta)
= \text{argmax}\;\prod_n^N P((S,R,T)_n|\Theta) \end{equation}

Considering the log-likelihood at a single training example $(S, R, T)$ and
taking the derivative with respect to parameters, we obtain:
\begin{equation}
\begin{split} \frac{\partial \log P(S, R, T|\Theta)}{\partial \Theta_i}
= &\frac{\partial}{\partial \Theta_{i}}\left[
-\mathcal{E}(S,R,T|\Theta)\right]\\ & -\frac{\partial}{\partial
\Theta_{i}}\left[ \log \sum_{s,r, t}e^{-\mathcal{E}(S, R, T | \Theta)}\right]
\end{split} \label{eq:contrastive} \end{equation}

Given a smooth energy function the first term is easily obtained, but the
second term is problematic. This term, derived from the partition function
$Z(\Theta)$, is intractable to evaluate in practice owing to its double sum
over the size of the vocabulary (potentially $\mathcal{O}(10^5)$).  In order to
circumvent this intractability we resort to techniques used to train Restricted
Boltzmann Machines and use stochastic maximum likelihood, also known as
persistent contrastive divergence (PCD); \cite{tieleman2008training}. In
contrastive divergence, the gradient of the partition function is estimated
using samples drawn from the model distribution seeded at the current training
example \cite{hinton2002training}. However, many rounds of sampling may be
required to obtain good samples. PCD retains a persistent Markov chain of model
samples across gradient evaluations, assuming that the underlying distribution
changes slowly enough to allow the Markov chain chain to mix. We use Gibbs
sampling by iteratively using the conditional distributions of all variables
($S$, $R$, and $T$, see below) to obtain model samples.

In particular, we draw $S$, $R$ and $T$ from the conditional probability
distributions:
\begin{equation} \begin{split} P(S| r, t; \Theta)
& =  \frac{e^{-\mathcal{E}(S, r, t|\Theta)}}{\sum_{s'}e^{-\mathcal{E}(s', r,
t|\Theta)}}\\ P(R | s, t; \Theta) & =\frac{e^{-\mathcal{E}(s, R,
t|\Theta)}}{\sum_{r'}e^{-\mathcal{E}(s, r', t|\Theta)}}\\ P(T |s, r; \Theta)
& =\frac{e^{-\mathcal{E}(s, r, T|\Theta)}}{\sum_{t'}e^{-\mathcal{E}(s, r,
t'|\Theta)}} \label{eq:conditionals} \end{split} \end{equation}
Thereby, we can estimate the gradient of $Z(\Theta)$ at the cost of these
evaluations, which are linear in the size of the vocabulary.

Using this, following the objective from (\ref{eq:contrastive}) further
simplifies to a contrastive objective given a batch of $B$ data samples and $M$
model samples (each model sample obtained from an independent, persistent
Markov chain): 
\begin{equation} \begin{split} \frac{\partial
P(\mathcal{D}|\Theta)}{\partial \Theta_i} &\simeq
\frac{1}{M}\sum_{m=1}^M\left[\frac{\partial
\mathcal{E}((S,R,T)_m|\Theta)}{\partial \Theta_i}\right]\\ &-
\frac{1}{B}\sum_{b=1}^B\left[\frac{\partial\mathcal{E}((S,R,T)_b|\Theta)}{\partial
\Theta_i}\right] \end{split} \end{equation}

Interestingly, the model can gracefully deal with missing elements in observed
triplets (for instance missing observed relationships). Learning is achieved by
considering the partially observed triple as a superposition of all possible
completions of that triple, each weighted by its conditional probability given
the observed elements, using (\ref{eq:conditionals}). This produces a
gradient which is a weighted sum.

In the fully-observed case (which we
sometimes call supervised in an abuse of terminology), the weighting is simply
a spike on the observed state. Similarly, the model can predict missing values 
as a simple inference step. These properties make having a joint distribution
very attractive in practical use,  offsetting the conceptual difficulty of 
training. In our experiments, we exploit these properties to do principled 
semi-supervised and unsupervised learning with partially observed or unobserved 
relationships without needing an external noise distribution or further 
assumptions.

\paragraph{Implementation}
We provide the algorithm in Python (\url{https://github.com/corcra/bf2}).
Since most of its runtime takes place in vector operations, we are developing
a GPU-optimised version.
We use \texttt{Adam} \cite{kingma2014adam} to adapt learning rates and improve
numerical stability. We used the recommended hyperparameters from this paper;
$\lambda = 1 -10^{-8}$, $\epsilon = 1-10^{-8}$, $\beta_{1} = 0.9$, $\beta_{2}
= 0.999$. Unless otherwise stated, hyperparameters specific to our model were:
dimension $d=100$, batch size of $B=100$, learning rate for all parameter types of $\alpha = 0.001$, and three rounds of Gibbs sampling to obtain model samples.

\section{Experiments}
We will proceed to explore the model in five
settings. First, an entity vector embedding problem on WordNet which 
consists of fully observed triplets of words and relationships. In the second 
case we demonstrate the power of the semi-supervised extension of the algorithm
on the same task.  We then show that a) adding relationship data can lead to 
better embeddings and b) that adding unstructured text can lead to better 
relationship predictions.  Finally, we demonstrate that the algorithm can also 
identify latent relationships that lead to better word embeddings.

\paragraph{Data} As structured data, we use the WordNet dataset
described by \citetext{socher2013reasoning}, available at
\url{http://stanford.io/1IENOYH}.  This contains 38,588 words and 11 types of
relationships. Training data consists of true triples such as
(\underline{feeling}, \underline{has instance}, \underline{pride}).

We derived an additional version of this dataset by stripping sense IDs from
the words, which reduced the vocabulary to 33,330 words. We note that this
procedure likely makes prediction on this data more difficult, as every word
receives only one representation. We did this in order to produce an aligned 
vocabulary with our \emph{unstructured} data source, taken to be English 
Wikipedia (\url{https://dumps.wikimedia.org/}, August 2014). We extracted text using
\texttt{WikiExtractor} (\url{http://bit.ly/1Imz1WJ}). We greedily identified
WordNet 2-grams in the Wikipedia text. Two words were considered in
a \underline{sentence context} if they appeared within a five word window.
Only pairs for which both words appeared in the WordNet vocabulary
were included.  We drew from a pool of 112,581 training triples in
WordNet with 11 relationships, and 8,206,304 triples from Wikipedia
(heavily sub-sampled, see experiments). To check that our choice to strip
sense IDs was valid, we also created a version of the Wikipedia dataset where 
each word was tagged with its most common sense from the WordNet 
training corpus. We found that this did not significantly impact our results,
so we chose to continue with the sense-stripped version, preferring to collapse
some WordNet identities over assigning possibly-incorrect senses to
words in Wikipedia.

\paragraph{WordNet Prediction Task} We used our model to solve the basic
prediction task described in \citetext{socher2013reasoning}. In this case, the
model must differentiate true and false triples, where false triples are
obtained by corrupting the $T$ entry in the triple, e.g. $(S, R, T) \rightarrow (S, R, \tilde{T})$ (where $(S, R, \tilde{T})$ doesn't appear in the training data). The `truth' of a triple is evaluated by its energy
$\mathcal{E}(S, R, T)$, with a relationship-specific cut-off chosen by
maximizing accuracy on a validation set (this is an equivalent
procedure to the task as initially described). By learning explicit
representations of each of the 38,588 entities in WordNet, our approach
most closely follows the \emph{`Entity Vector'} task in Socher \emph{et al.}
This is to be contrasted with the \emph{`Word Vector'} task, where a
representation is learned for each word, and entity representations are
obtained by averaging their word vectors. We elected not to perform this task
because we are not confident that composition into phrases through averaging
is well-justified. Using the validation set to select an early stopping point at 
66 epochs, we obtain a test set accuracy of $78.2\%$ with an AUROC of $85.6\%$.
The `Neural Tensor Model' (NTN) described in \citetext{socher2013reasoning} achieves an
accuracy of around $70\%$ on this task, although we note that the simpler
Bilinear model also described in \citetext{socher2013reasoning} achieves $74\%$
and is closer to the energy function we employ. The improved
performance exhibited by this simpler Bilinear model was also noted by
\citetext{export:241703}. Other baselines reported by Socher \emph{et al.} were a single layer model without an interaction term, a Hadamard model \cite{bordes2012joint} and the model of \citetext{bordes2011learning} which learns separate left and right relationship operators for each element of the triple. These were outperformed by the Bilinear and NTN models, see Figure 4 in \citetext{socher2013reasoning} for further details. Hence, our model outperforms the two previous methods by more than $4\%$.

As a preliminary test of our model, we also considered the \texttt{FreeBase}
task described by \citetext{socher2013reasoning}. Initial testing yielded an
accuracy of 85.7\%, which is comparable to the result of their best-performing
model (NTN) of about 87\%. We chose not to further explore this dataset
however, because its entities are mostly proper nouns and thus seemed unlikely
to benefit from additional semantic data.

\begin{figure}[h]
$ $\hspace*{3ex}\includegraphics[width=0.4\textwidth]{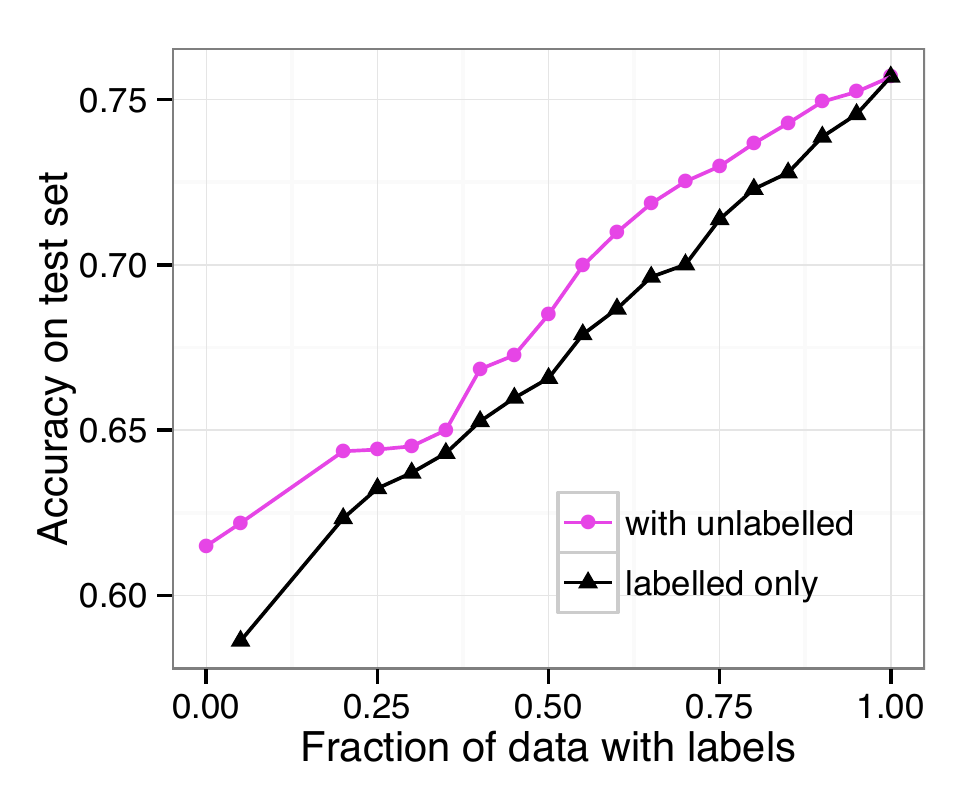}
\caption{\textbf{Semi-supervised learning improves learned embeddings:} We
tested the semi-supervised extension of our approach on the entity relationship
learning task described in \citetext{socher2013reasoning} and previous subsection.
Following Socher \emph{et al.}, we predict if a triple $(S, R, T)$ is true by
using its energy as a score. For this we trained our algorithm with a subset of
the training data (total 112,581 examples). The fully-supervised version only used the
subset of fully labelled data (varying amounts as indicated on the x-axis).
For semi-supervised learning, in addition we use the remaining data but where
the type of the relationship between pairs is unknown. We find that the
semi-supervised method consistently performs better than the fully supervised
method (see main text for more details).  \label{fig:figB}}
\end{figure}

\paragraph{Semi-supervised Learning for WordNet} We next tested the
semi-supervised learning capabilities of our algorithm (see Inference and
Learning). For this we consider the same task as before, but omit some label
information in the training set and instead use posterior probabilities during
the inference.  For this we trained our algorithm with a subset of the training
data (total 112,581 examples) and measured the accuracy of classifying into
true and false relationships as before. The fully-observed case used only
a subset of fully-observed data (varying amounts as indicated on the x-axis).
For semi-supervised learning, we also used the remaining data, but
masking the type of the relationship between pairs. In
\textbf{Figure~\ref{fig:figB}} we report the accuracy for different
labelled/unlabelled fractions of otherwise the same dataset.  We find that the
semi-supervised method consistently performs better than the fully observed
method for all analysed training set sizes. In this and the previous experiment,
one Markov chain was used for PCD
and a $l_2$ regulariser on $G_R$ parameters
with weight 0.01.

\begin{figure}[h]
\includegraphics[width=0.27\textwidth]{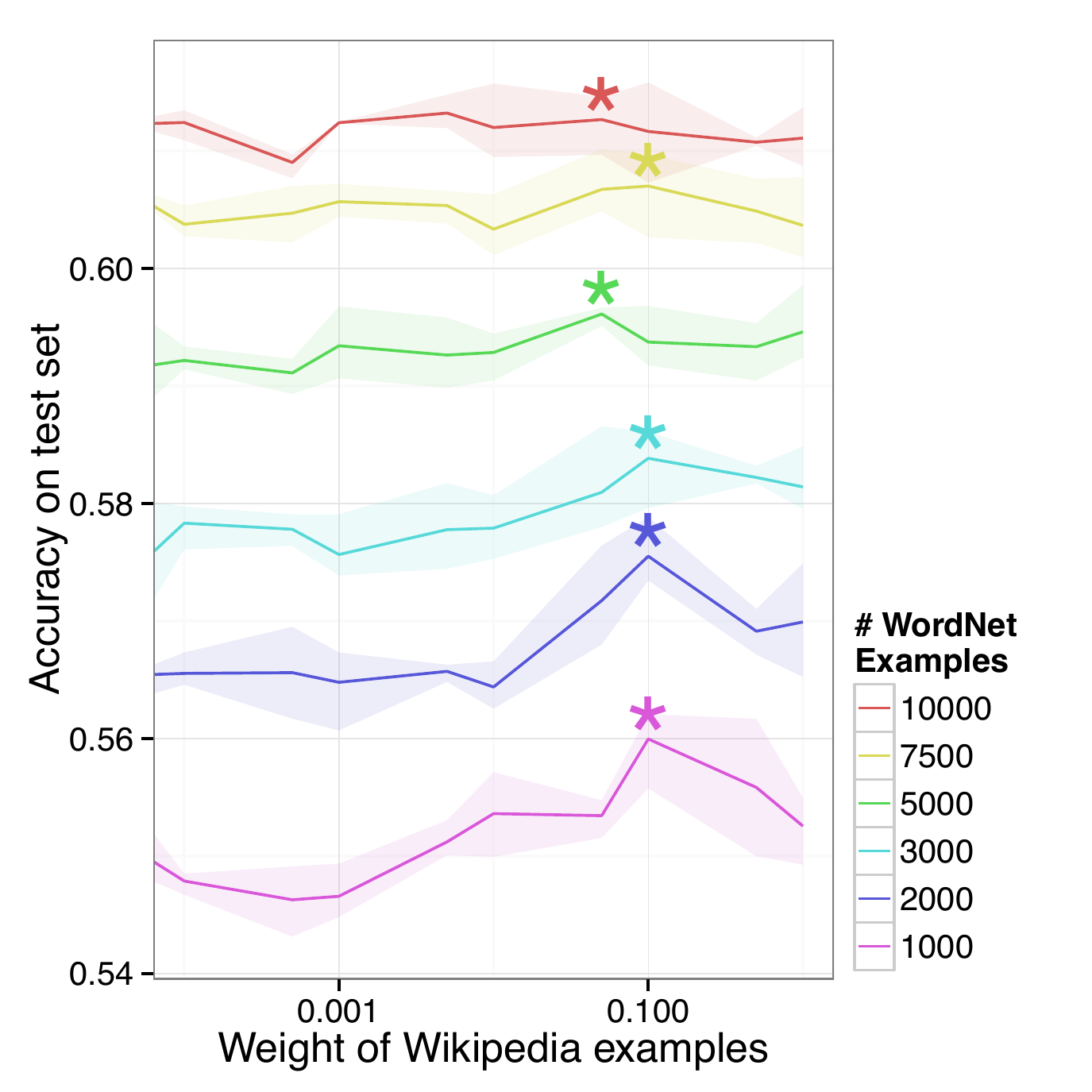}\includegraphics[width=0.19\textwidth]{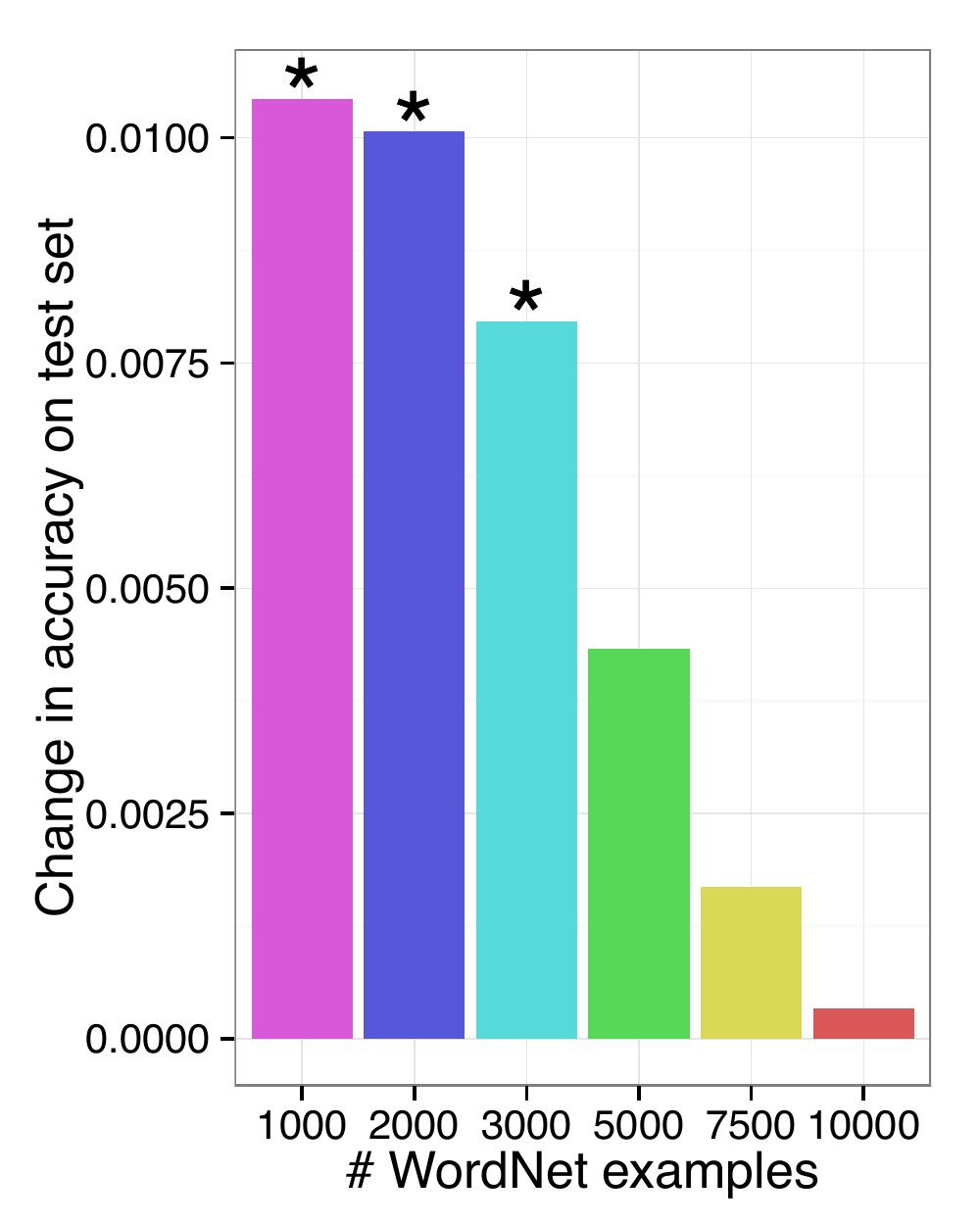}

\caption{\textbf{Unstructured data helps relationship learning:} In addition to
training on a set of known relationships, we use unstructured data from
Wikipedia with varying weight ($x$-axis) during training. As before,
with the goal to predict if a triple $(S, R, T)$ is true by using its energy as
a score. A validation set is used to determine the threshold below which
a triple is considered `true'. The solid line denotes the average of
three independent experimental runs; shaded areas show the range of results. 
The bar plot on the right shows the difference in accuracy
between $\kappa = 0$ and $\kappa = \kappa^{*}$, where $\kappa^{*}$ gave the
highest accuracy on a validation set. Significance at 5\% (paired t-test) is
marked by asterisk. We find then that unstructured Wikipedia can improve relationship 
learning in cases when labelled relationship data is scarce.
\label{fig:transfer}
} \end{figure}

\paragraph{Adding Unstructured Data to a Relationship Prediction Task} To test
how unstructured text data may improve a prediction task when \emph{structured}
data is scarce, we augmented a subsampled set of triples from WordNet
with 10,000 examples from Wikipedia and varied the weight $\kappa$ associated
with their
gradients during learning. The task is then to predict whether or not a given triple
$(S, R, T)$ is a true example from WordNet, as described previously.
\textbf{Figure~\ref{fig:transfer}} shows accuracy on this task as $\kappa$ and
the amount of structured data vary. To find the improvement associated with
\emph{unstructured data}, we compared accuracy at $\kappa = 0$ with $\kappa
= \kappa^{*}$ (where $\kappa^{*}$ gave the highest accuracy on the validation
set; marked with $\ast$).  We find that including free text data quite
consistently improves the classification accuracy, particularly when structured data is scarce.

In this experiment and all following, we used five Markov chains for PCD and
a $l_2$ regulariser on all parameters with weight 0.001.

\paragraph{Relationship Data for Improved Embeddings} In this case, we assume
\emph{unstructured text data} is restricted, and vary the quantity of
structured data. To evaluate the \emph{untransformed} embeddings, we use them
as the inputs to a supervised multi-class classifier. The task for a given $(S,
R, T)$ triple is to predict $R$ given the vector formed by concatenating
$\mathbf{c}_{S}$ and $\mathbf{v}_{T}$. We use a random forest classifier
trained on the WordNet validation set using five-fold cross-validation.

\begin{figure}[h] \noindent$
$\hspace*{-2ex}\includegraphics[width=0.49\textwidth]{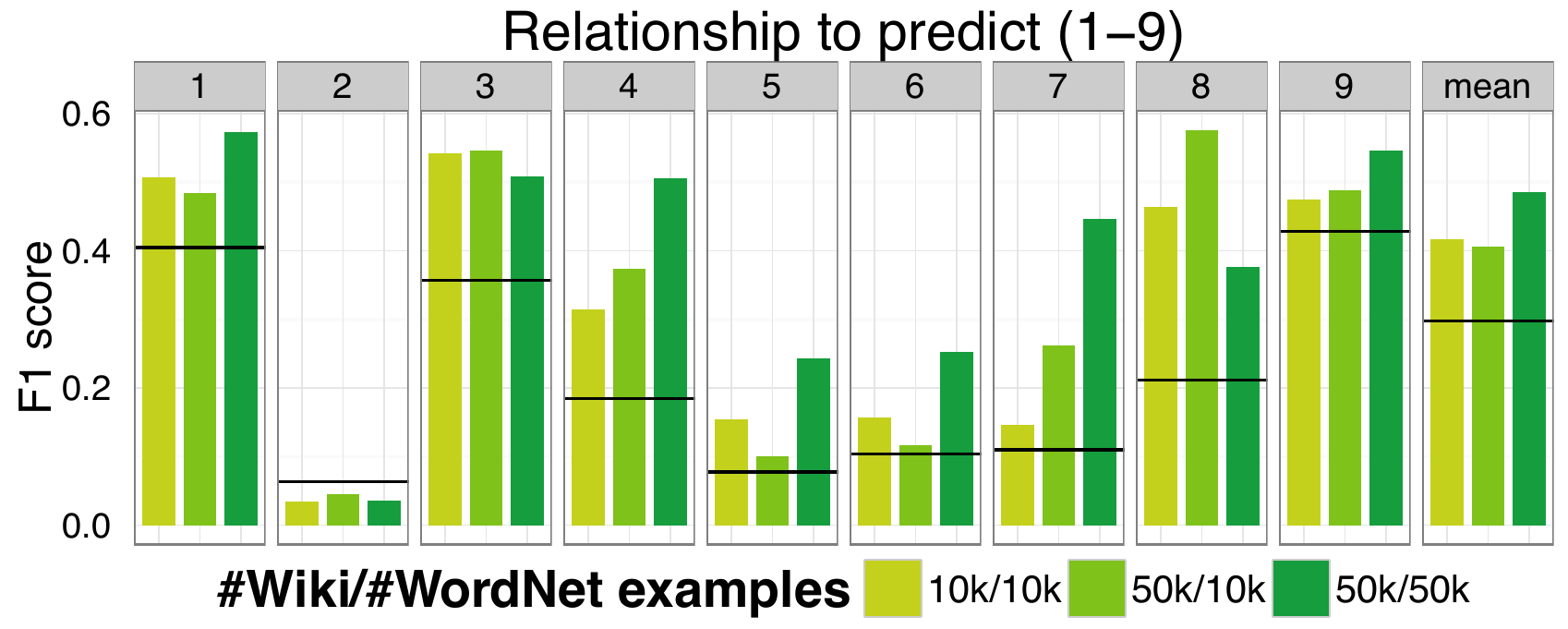}
\caption{\textbf{Relationship data improves learned embeddings:} We apply our
algorithm on a scarce set of Wikipedia co-occurences (10k and 50k instances)
with varying amounts of additional, unrelated relationship data (10k and 50k
relations from WordNet). We test the quality of the embedding by
measuring the accuracy on a task related to nine relationships (\underline{has
instance}, \underline{domain region}, \underline{subordinate instance of},
\underline{member holonym}, \underline{has part}, \underline{has part},
\underline{part of}, \underline{member meronym}, \underline{synset domain
topic}, \underline{type of}; relationships \underline{similar to},
\underline{domain topic} were omitted for technical reasons). We used eight of
the relationships together with the Wikipedia data to learn representations
that are then used in a subsequent supervised learning task to predict the
remaining ninth relationship based on the representations using random forests.
Black lines denote results from \texttt{word2vec} trained on a Wikipedia-only
dataset with 4,145,372 \emph{sentences}. \label{fig:embedding}\vspace*{-2.5ex}} \end{figure}

To avoid testing on the training data (since the embeddings are obtained using
the WordNet training set), we perform this procedure once for each
relationship (11 times - excluding \underline{appears in sentence with}), each time removing from the training data \emph{all}
triples containing that relationship. \textbf{Figure~\ref{fig:embedding}} shows
the $F1$ score of the multi-class classifier on the left-out relationship for
different combinations of data set sizes. We see that for most relationships,
including more unstructured data improves the embeddings (measured by
performance on this task). We also trained \texttt{word2vec} \cite{mikolov2013distributed} on a much larger Wikipedia-only dataset (4,145,372 \emph{sentences}) and trained a classifier on its vectors; results are shown as black lines. 
We see that our approach yields a consistently higher $F1$ score, suggesting that even data about \emph{unrelated} relationships provides information to produce vectors that are semantically richer overall.

\emph{These results illustrate that embeddings learned from limited free text
data can be improved by additional, unrelated relationship data.}

\begin{figure}[h]
\hspace*{-2ex}\includegraphics[width=0.49\textwidth]{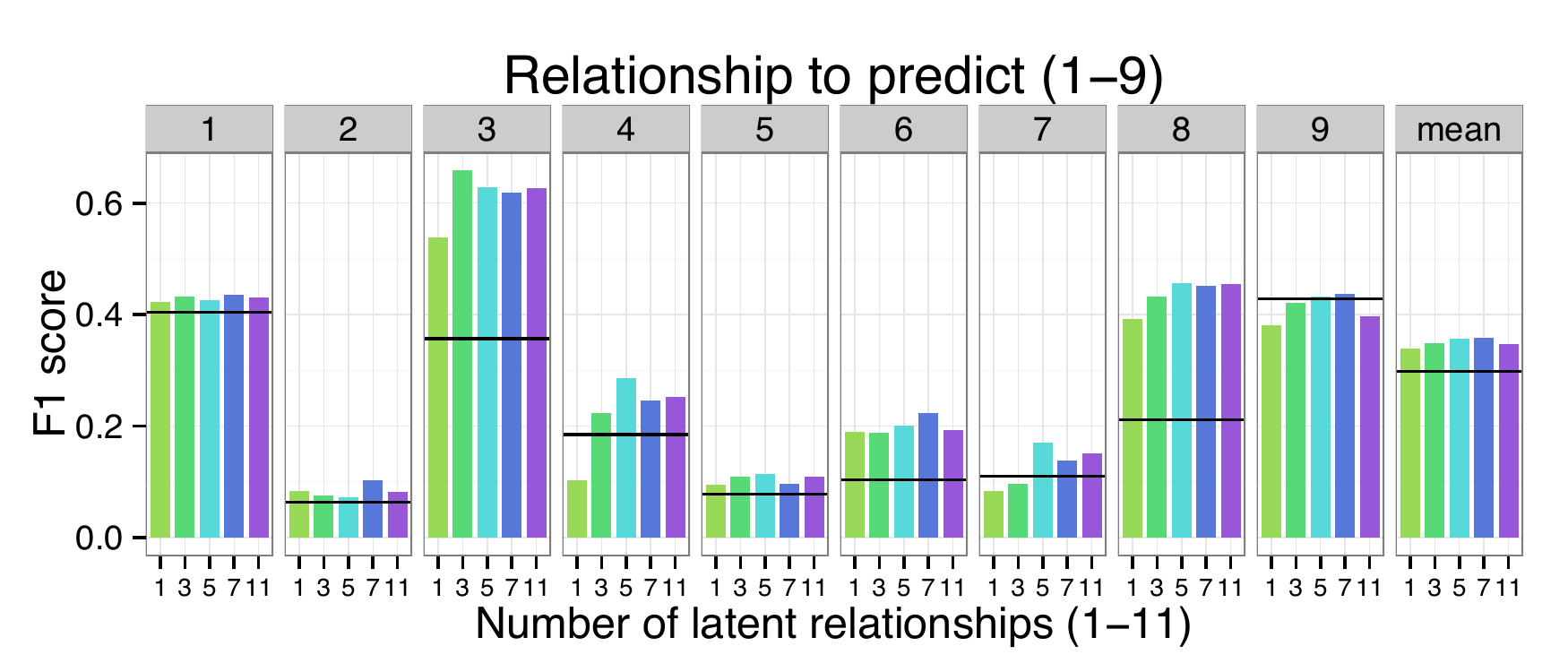}
\caption{\textbf{Unsupervised learning of latent relationships improves
embeddings}: We train a fully unsupervised algorithm with 1, 3, 5, 7 and 11
possible latent relationships on one million Wikipedia sentences. Initialisation
is at random and without prior knowledge.  To test the quality of the resulting
\emph{embeddings}, we use supervised learning of nine WordNet
relationships with random forests. Depending on the relationship at hand, the
use of multiple latent relationships during training leads to slightly, but
consistently better accuracies using the computed embeddings for every of the
nine relationships and also on average.  Hence, the resulting embeddings using
unsupervisedly learned latent relationships can be said to be of higher
quality. Once again, black lines show results using \texttt{word2vec}.
\label{fig:figF}} \end{figure}

\paragraph{Unsupervised Learning Of Relationships} In our final
experiment, we explore the ability of the model to learn embeddings from
co-occurrence data alone, without specifying the relationships it should use.
When using the model with just one relationship (trivially the identity), the
model effectively reverts to \texttt{word2vec}. However, if we add a budget of
relationships (in our experiments we use 1, 3, 5, 7, 11), the model has additional
parameters available to learn affine transformations of the space which can
differentiate how distances and meaning interact for the word embeddings
without fixing this \emph{a priori}. Our intuition is that we want to test
whether textual context alone has substructure that we can capture with latent
variables. We generate a training set of one million word co-occurrences from
Wikipedia (using a window size of 5 and restricting to words appearing
in the WordNet dataset, as described earlier), and train different
models for each number of latent relationships. Inspired by earlier experiments
testing the utility of supplanting WordNet training data with
Wikipedia examples, we decide to test the ability of a model purely
trained on Wikipedia to learn word and relationship representations
which are predictive of WordNet triplets, \emph{without} having seen
any data from WordNet. As a baseline we start with
$|R|=1$ to test how well word embeddings from context alone can perform, indicated
by the leftmost bar in \textbf{Figure~\ref{fig:figF}}. We then proceed to
train models with more latent relationships. We observe that, especially for
some relationship prediction tasks, including this flexibility in the model
produces a noticeable increase in $F1$ score on this task. Since we evaluate
the embeddings alone, this effect must be due to a shift in the content of
these vectors, and cannot be explained by the additional parameters introduced
by the latent relationships. {\it We note that a consistent explanation for
this phenomenon is that the model discovers contextual subclasses which are
indicative of WordNet-type relationships. This observation opens doors
to further explorations of the hypothesis regarding contextual subclasses and
unsupervised relationships learning from different types of co-occurrence
data}.
  
We note that we did not perform an exhaustive search of the hyperparameter 
space; better settings may yet exist and will be sought in future work. Nonetheless, although the absolute improvement in $F1$ score yielded by this method is modest, we are encouraged by the model's ability to exploit latent variables in this way.

\section{Discussion} We have presented a \emph{probabilistic generative model
of words and relationships} between them.  By estimating the parameters of this
model through stochastic gradient descent, we obtain vector and matrix
representations of these words and relationships respectively. To make learning
tractable, we use \emph{persistent contrastive divergence} with Gibbs sampling
between entity types ($S$, $R$, $T$) to approximate gradients of the partition
function.  Our model uses an energy function which contains the idealised
\texttt{word2vec} model as a special case. By augmenting the embedding space
and considering relationships as arbitrary \emph{affine} transformations, we
combine benefits of previous models. In addition, our formulation as
a generative model is distinct and allows a more flexible use, especially in
the missing data, semi- and unsupervised setting.  Motivated by domain-settings
in which structured \emph{or} unstructured data may be scarce, we illustrated
how a model that combines both data sources can improve the quality of
embeddings, supporting other findings in this direction.

A promising field of exploration for future work is a more detailed treatment of
relationships, perhaps generalising from affine transformations to include
nonlinear maps. Our choice of cosine similarity in the energy function can also
be developed, as we note that this function is insensitive to very small
deviations in angle, and may therefore produce looser clusters of synonyms.
Nonlinearity could also be introduced in the energy, using for example splines.
Furthermore, we intend to encode the capacity for richer transfer of structured
information from sources such as graphs as prior knowledge into the model. Our
current model can take advantage of local properties of graphs to that purpose,
but has no explicit encoding for nested and distributed relationships.

A limitation of our model is its conceptual inability to embed whole sentences
(which has been tackled by averaging vectors in other work, but requires deeper
investigation). Recurrent or more complex neural language models offer many avenues to pursue as extensions for our model to tackle this.  A particularly interesting direction to achieve that would be a combination with work such as~\cite{jernitefast},
which could in principle be integrated with our model to include relationships.

The intended future application of this model is exploratory semantic data
analysis in domain-specific pools of knowledge. We can do so by combining prior
knowledge with unstructured information to infer, for example, new edges in
knowledge graphs. A promising such field is \emph{medical language processing},
retrospective exploratory data analysis may boost our understanding of the 
complex relational mechanisms inherent in multimodal observations, and specific
medical knowledge in the form of (for example) the UMLS can be used as a strong 
regulariser. Indeed, initial experiments combining clinical
text notes with relational data between UMLS concepts from SemMedDB \cite{Kilicoglu01122012} have
demonsrated the utility of this combined approach to predict the functional
relationship between medical concepts, for example, \texttt{cisplatin} \underline{\texttt{treats}} \texttt{diabetes}. We are in the process of expanding this investigation.

\section{Acknowledgments} This work was funded by the Memorial Hospital and the
Sloan Kettering Institute (MSKCC; to G.R.). Additional support for S.L.H. was
provided by the Tri-Institutional Training Program in Computational Biology and
Medicine.
%%%%%%%%%%
% References and End of Paper
\bibliography{hyland_references}
\bibliographystyle{aaai} \end{document}